\documentclass[lettersize,journal]{IEEEtran}
\usepackage{amsmath,amsfonts}
\usepackage{array}
\usepackage[caption=false,font=normalsize,labelfont=sf,textfont=sf]{subfig}
\usepackage{textcomp}
\usepackage{stfloats}
\usepackage{url}
\usepackage{verbatim}
\usepackage{graphicx}
\usepackage{cite}
\usepackage{booktabs}
\usepackage{xcolor}
\usepackage{amssymb}
\usepackage{makecell} 
\usepackage[ruled,linesnumbered]{algorithm2e}

\usepackage[table]{xcolor} 
\definecolor{rowcolor}{rgb}{0.898, 0.949, 0.969}
\usepackage{multirow} 
\usepackage{mathrsfs}
\usepackage[colorlinks,
linkcolor=blue,
anchorcolor=black,
citecolor=black]{hyperref}
\usepackage{bm}
\usepackage[font=footnotesize, skip=4pt]{caption}
\begin{document}

\title{From Mannequin to Human: A Pose-Aware and Identity-Preserving Video Generation Framework for Lifelike Clothing Display}

\author{Xiangyu Mu, Dongliang Zhou, Jie Hou, Haijun Zhang, Weili Guan

\thanks{X. Mu, D. Zhou, J. Hou, H. Zhang, and W. Guan are with Harbin Institute of Technology, Shenzhen, Xili University Town, Shenzhen 518055, China. 
Corresponding authors: Dongliang Zhou (e-mail: zhoudongliang@hit.edu.cn) and Haijun Zhang (e-mail: hjzhang@hit.edu.cn).}
}



\maketitle

\begin{abstract}
Mannequin-based clothing displays offer a cost-effective alternative to real-model showcases for online fashion presentation, but lack realism and expressive detail. 
To overcome this limitation, we introduce a new task called mannequin-to-human (M2H) video generation, which aims to synthesize identity-controllable, photorealistic human videos from footage of mannequins.
We propose M2HVideo, a pose-aware and identity-preserving video generation framework that addresses two key challenges: the misalignment between head and body motion, and identity drift caused by temporal modeling. 
In particular, M2HVideo incorporates a dynamic pose-aware head encoder that fuses facial semantics with body pose to produce consistent identity embeddings across frames. To address the loss of fine facial details due to latent space compression, we introduce a mirror loss applied in pixel space through a denoising diffusion implicit model (DDIM)-based one-step denoising. Additionally, we design a distribution-aware adapter that aligns statistical distributions of identity and clothing features to enhance temporal coherence.
Extensive experiments on the UBC fashion dataset, our self-constructed ASOS dataset, and the newly collected MannequinVideos dataset captured on-site demonstrate that M2HVideo achieves superior performance in terms of clothing consistency, identity preservation, and video fidelity in comparison to state-of-the-art methods.
\end{abstract}


\section{Introduction}
\label{sec:intro}

\IEEEPARstart{T}{he} rapid development of social media platforms has significantly changed how people share content and conduct advertising online. This transformation has intensified the demand for high-quality, visually compelling digital experiences, particularly in domains where visual presentation directly influences consumer engagement. The fashion industry is a prime example, projected to generate \$880.91 billion in revenue in 2025 and projected to reach \$1.18 trillion by 2029.\footnote{\url{https://www.statista.com/outlook/emo/fashion/worldwide}}
Due to the massive size of the market and a growing focus on visual engagement, researchers have delved into various areas of fashion technology, including garment reconstruction \cite{hong2021garment4d,zhan2024pattern} and digital human visualization \cite{yu2023monohuman,panagiotidou2022communicating}.
Despite this progress, the problem of enhancing the visual realism and expressiveness of the presentation of the clothing remains comparatively underexplored in computer vision and graphics. 
Current modes of presentation typically fall into three categories:flat-lay images, mannequin displays, or real model showcases.
Among these options, real models are the most effective at communicating clothing fit, drape, and motion. However, producing content with real models requires substantial effort and financial investment, making it inaccessible for many small- and medium-sized retailers. In contrast, flat-lay and mannequin displays are more affordable but lack realism and expressive detail, reducing their effectiveness at engaging consumers. To bridge this gap, we propose a novel task, referred to as the mannequin-to-human (M2H) video generation task. This task transforms videos of mannequin-dressed clothing into photorealistic human representations. It not only preserves the appearance of the clothing with high fidelity but also allows explicit control over the identity of the generated human avatars.

\begin{figure*}[ht!]
\centering
\includegraphics[width=0.96\textwidth]{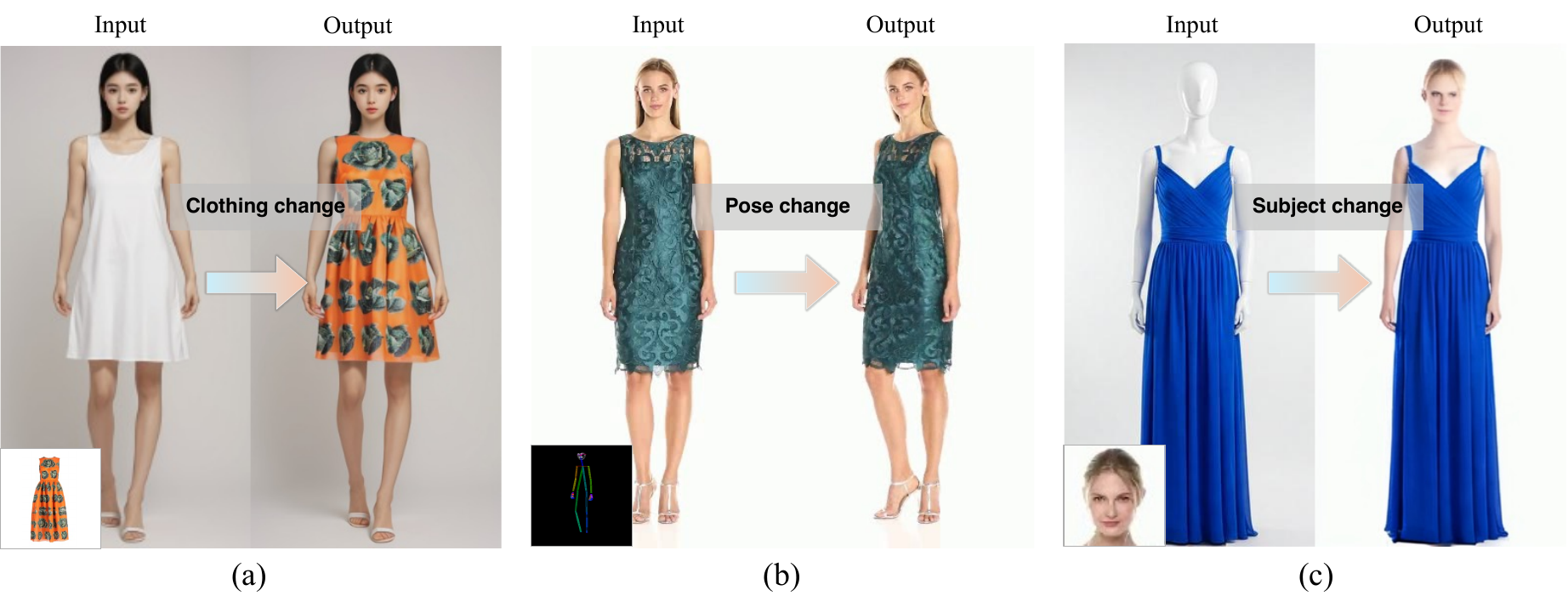}
\caption{Comparison of tasks related to M2H video generation. (a) Virtual try-on, which focuses on transferring clothing onto a target model while preserving identity. (b) Pose transfer, which aims to reproduce body posture while maintaining the fidelity of the appearance. (c) Mannequin-to-human video generation (ours), which generates photorealistic human frames that retain the clothing of the mannequin and adopt the features of the identity of the reference subject.}
\label{fig:comparision4diff-tasks}
\end{figure*}

The tasks most closely related to our proposed M2H video generation task are virtual try-on (VTON) \cite{morelli2023ladi,xu2025ootdiffusion,chong2025catvton,zhang2025viton} and pose transfer \cite{bhunia2023person,karras2023dreampose,lu2024coarse}, as illustrated in Figs. \ref{fig:comparision4diff-tasks}(a) and (b), respectively.
VTON focuses on replacing clothing in images or videos with target clothing while preserving the subject's original identity. 
Recent methods \cite{morelli2023ladi,xu2025ootdiffusion,chong2025catvton} have adopted
latent diffusion models (LDMs) \cite{rombach2022high,brooks2023instructpix2pix,gal2023image} to improve realism and controllability.
Initially, Morelli \textit{et al.} \cite{morelli2023ladi} applied LDMs to VTON. They proposed a text inversion module based on contrastive language-image pre-training (CLIP) \cite{radford2021learning} to encode clothing semantics into the generation process.
However, the compressed latent space in LDMs tends to lose high-frequency details.
To overcome this, Xu \textit{et al.} \cite{xu2025ootdiffusion} proposed a dual-UNet architecture, where the additional outfitting UNet learns clothing-specific details that naturally share the latent space with the backbone layers.
Although VTON has advanced, it cannot be directly adapted for the M2H video generation task due to some key differences. VTON remains within the human domain, while M2H requires cross-domain transformation from mannequins to photorealistic humans with identity control as shown in Fig. \ref{fig:comparision4diff-tasks}(c). This involves addressing spatial misalignment, maintaining facial consistency across frames, and incorporating the identity from a reference image, all beyond VTON's scope.
Pose transfer, depicted in Fig. \ref{fig:comparision4diff-tasks}(b), aims to generate realistic images or videos by transferring human poses while maintaining identity and appearance. 
LDMs have also been widely adopted for this task. Bhunia \textit{et al.} \cite{bhunia2023person} incorporated a texture encoder to extract multi-scale feature groups for improved image textures.
Subsequently, DreamPose \cite{karras2023dreampose} improves pose control and identity preservation through a dual-path encoder that combines semantics based on CLIP and textures based on the variational autoencoder (VAE) \cite{Kingma2013AutoEncodingVB}.
However, these methods overlook the complex interdependence between pose and appearance, which limits their ability to generalize effectively.
To address this, Xiao \textit{et al.} \cite{xiao2025disentangled} introduced a multi-pose generation framework that disentangles pose and appearance by a global-aware module and appearance adapter.
Although notable progress has been made, pose transfer methods remain ill-suited to the M2H video generation task. In pose transfer methods, the information about the identity, specifically the coordinates of the facial keypoints, is provided explicitly for each frame. In contrast, M2H requires synthesizing a temporally coherent identity based solely on a single reference image, which is often captured under unconstrained conditions.
In addition to the distinct characteristics that differentiate M2H video generation from both VTON and pose transfer, there are also critical common challenges that set M2H apart from these tasks.
When VTON and pose transfer tasks are extended from image generation to video generation,
existing works \cite{morelli2023ladi,lu2024coarse,xu2025ootdiffusion} typically add a temporal layer to the LDM backbone to improve temporal consistency.
However, applying temporal layers directly to the M2H video generation task may introduce additional challenges. In particular, temporal layers can cause inter-frame interference, wherein features at the same spatial location become temporally context-dependent. This results in shifts in the distribution statistics, such as shifts in the mean and variance from one frame to another. Consequently, critical regions such as faces may suffer from corruption or blurring, especially in scenarios involving rapid motion.
In summary, the M2H video generation task should address two issues: (i) the accurate prediction of facial pose based on body dynamics while preserving identity features, and (ii) the mitigation of shifts in the distribution and inter-frame interference.

To address these issues, we propose a new video generation framework, referred to as M2HVideo, designed to transform mannequin-based clothing displays into photorealistic human representations while preserving clothing fidelity and human identity. To tackle the first issue, we propose a dynamic pose-aware head encoder. It leverages both hybrid facial features and dynamic body pose information. It produces an identity embedding capable of accurately estimating the facial position even in the presence of complex body motion. The hybrid facial features include semantic facial features extracted by ArcFace \cite{deng2019arcface} and spatial representations derived from a VAE encoder. The generated identity embedding is integrated into the denoising UNet of the LDM via an adapter, guiding the generation of the facial region in each video frame.
However, relying exclusively on latent space operations can result in the loss of high-frequency details.
Our preliminary experiments confirmed that LDMs may fail to preserve fine facial characteristics. These include details such as wrinkles and skin texture, particularly when such features occupy only a small region of the entire image.
According to Rate-Distortion Theory \cite{shannon1959coding}, this degradation results from the encoder of LDMs compressing high-dimensional images into low-dimensional latent codes.
Existing methods \cite{morelli2023ladi,lu2024coarse,xu2025ootdiffusion} commonly apply loss functions in the compressed latent space. However, this strategy limits the expressive capacity of the LDMs, as their performance becomes constrained by the representational bounds of the latent space, consistent with the theory of the information bottleneck \cite{tishby2015deep}.
To alleviate this phenomenon, we develop a mirror loss that operates in the pixel space. In particular, inspired by the denoising diffusion implicit model (DDIM) \cite{song2020denoising}, we approximate the denoised latent code at step $0$ from an intermediate latent code at step $t$, using the predicted noise and its corresponding intensity.
The approximated latent code at the initial step is then passed through the LDM decoder to reconstruct the image in the pixel space. 
The mirror loss is then applied to reduce the discrepancy in the facial region between the generated output and the corresponding ground-truth video frames.
This strategy can enhance the preservation of high-frequency details.
For the second issue, we introduce a distribution-aware adapter. This module performs two separate cross-attention operations: one between the latent code and the CLIP-derived clothing embedding, and the other between the latent code and the identity embedding obtained from the dynamic pose-aware head encoder.
The outputs of these two cross-attention branches are modulated by aligning their respective distributions.
In particular, the identity features are first normalized using their mean and standard deviation. They are then rescaled to match the mean and standard deviation of the clothing features.
This modulation reduces distributional shifts in the latent code and stabilizes the feature representation across video frames.
The proposed framework is evaluated on the UBC fashion dataset \cite{zablotskaia2019dwnet}, a self-constructed ASOS dataset, and a new MannequinVideos dataset made up of real-world mannequin recordings.
These datasets are available at \url{https://huggingface.co/datasets/MML-Group/M2HVideo-data}.
Extensive experiments show that our proposed framework outperforms several state-of-the-art methods in terms of various evaluation metrics.

The main contributions of this research can be summarized as follows:
\begin{itemize}
\item To the best of our knowledge, this is the first framework that transforms mannequin-based clothing displays into photorealistic human videos, while preserving both the consistency of the clothing and the identity. 
\item 
We introduce a dynamic pose-aware head encoder that leverages both hybrid facial features and dynamic body pose to generate identity embeddings. This encoder is primarily designed to guide the spatial alignment between the movements of the face and body. It uses a combination of facial appearance cues and body pose representations to ensure consistent and natural face positioning in accordance with dynamic gestures.
\item 
We design a distribution-aware adapter that modulates cross-attention outputs by aligning the statistical distributions of the identity and clothing features. This module is mainly applied to mitigate distributional shifts between appearance attributes and enhance the temporal consistency of the generated video.
\item 
We propose a mirror loss applied in the pixel space via DDIM-based one-step denoising to mitigate the loss of fine facial details caused by latent space compression. This loss function is primarily designed to be used to enhance the quality of the reconstruction of the facial features. It leverages the deterministic nature of DDIM to refine pixel-level outputs and recover high-frequency details more effectively.
\end{itemize}

The remainder of this paper is organized as follows. Section \ref{related_work} presents a comprehensive review of related work. The proposed M2HVideo framework is detailed in Section \ref{method}. Section \ref{experiment} describes the experimental setup and provides extensive quantitative and qualitative results. Finally, conclusions are drawn in Section \ref{conclusion}.

\section{Related Work}
\label{related_work}
Our work is most related to three streams of research: video-to-video (V2V) translation, virtual try-on, and pose transfer. In this section, we first briefly review the literature in these three realms. Then, we highlight the distinctive features of our approach in comparison to the existing methods.

\textbf{Video-to-Video Translation.}
Video-to-video translation transforms a video in the source domain into a target domain by modifying its appearance or style while preserving structure and temporal coherence. As a temporal extension of image-to-image translation, V2V translation faces additional challenges, including ensuring temporal consistency and semantic fidelity. To address these, Wei \textit{et al.} \cite{wei2018video} developed a framework based on generative adversarial networks (GANs) \cite{goodfellow2020generative} that employs a two-stream discriminator to enforce temporal consistency across video frames. Given the challenges in constructing paired training data for V2V tasks, Bashkirova \textit{et al.} \cite{bashkirova2018unsupervised} used an unsupervised method based on a 3D CycleGAN model. By treating videos as spatio-temporal tensors, their model effectively maintains the consistency of both the motion and the appearance across frames. Subsequently, world-consistent V2V \cite{mallya2020world} was proposed to enhance the long-range temporal coherence. With diffusion models, Yang \textit{et al.} \cite{yang2023rerender} proposed a zero-shot, text-guided LDM that ensures temporal consistency by generating key frames and propagating them via temporal-aware patch matching. 
These methods convert entire videos from one domain to another without retaining specific regions or fine details.
In contrast, M2H video generation preserves the clothing region. Additionally, while V2V translation relies on low-frequency semantic cues such as style, M2H requires the retention of high-frequency details in the guiding signals, including the facial features and wrinkles associated with a specific identity.

\textbf{Virtual Try-on.}
Recent research on VTON can be broadly categorized into image-based and video-based approaches. Among these, video-based VTON is most relevant to our task, as it operates directly on video sequences. 
Due to the powerful generative priors of pre-trained LDMs, recent methods adopt them as the backbone for high-quality results.
Initially, Fang \textit{et al.} \cite{fang2024vivid} introduced LDMs into this task. It uses a clothing encoder to extract detailed clothing semantics and employs a lightweight pose encoder to capture motion features.
To address flickering artifacts caused by complex poses, Karras \textit{et al.} \cite{karras2024fashion} introduced 3D convolutions and temporal attention blocks to enhance the temporal consistency. 
Furthermore, Xu \textit{et al.} \cite{xu2024tunnel} proposed isolating the clothing region and applying Kalman filtering for temporal smoothing. The smoothed features are then injected into attention layers.
Similarly, RealVVT \cite{li2025realvvt} developed an agnostic mask-guided attention loss to improve the spatial consistency within the clothing region. However, all of these methods operate exclusively within the human domain. Therefore, they cannot be directly applied to the M2H video generation task, which requires cross-domain transformation from mannequins to photorealistic humans while preserving identity.

\textbf{Pose Transfer.}
In contrast to image-based methods in pose transfer, video-based ones need to model the impact of the pose on the human appearance while maintaining temporal smoothness to prevent flickering artifacts. 
To investigate the potential of image-based LDMs for generating temporally coherent videos, Karras \textit{et al.} \cite{karras2023dreampose} employed VAE and CLIP encoders to extract appearance features jointly. They trained an LDM with continuous pose sequences to enhance the smoothness of the motion and the temporal consistency. 
Subsequent studies \cite{hu2024animate,lu2024coarse,gan2025humandit} commonly incorporated temporal layers into the LDMs to generate the video.
For instance, AnimateAnyone \cite{hu2024animate} uses
a symmetric UNet architecture with temporal attention to ensure visual consistency across frames. To address the distortion of the appearance caused by changes in the pose, Lu \textit{et al.} \cite{lu2024coarse} introduced a perception-refined decoder and a hybrid-granularity attention module. 
These components enable coarse-to-fine control, enhancing the quality of the image and the generalizability while preventing overfitting.
Furthermore, HumanDiT \cite{gan2025humandit} employed a diffusion transformer with keypoint-DiT to generate sequences of poses and a pose adapter for rendering fine details.
All these studies rely on frame-wise explicit supervision based on facial key points, whereas M2H is required to generate a temporally consistent identity using only a single reference image.

\textbf{Positioning of Our Work.}
Among the various V2V translation tasks explored in prior work \cite{wei2018video,bashkirova2018unsupervised,mallya2020world},
our task is unique in that it targets a specialized mannequin-to-human video translation setting. In particular, given a mannequin video dressed in the target clothing and a reference identity image, our goal is to synthesize a photorealistic human video that preserves the clothing and motion of the mannequin while transferring the identity from the reference image.
Unlike VTON methods \cite{fang2024vivid,karras2024fashion,xu2024tunnel} and pose transfer approaches \cite{karras2023dreampose,lu2024coarse,gan2025humandit}, which operate entirely within the human domain, our task requires cross-domain translation from mannequin to human. This introduces distinct challenges related to spatial misalignment, identity control, and facial synthesis. Therefore, unlike the aforementioned methods, our approach performs cross-domain conversion with an emphasis on preserving identity, requiring the removal of mannequin-specific textures and precise control over facial synthesis.

\section{M2HVideo}
\label{method}
In this section, we begin with a concise overview of the latent diffusion model and its extension to video generation, highlighting key definitions and strategies used in prior LDM-based approaches in Section~\ref{method:sec3.1}.
Then the formal definition of the problem of the proposed M2HVideo task is presented in Section~\ref{method:sec3.2}.
Next, Section~\ref{method:sec3.3} introduces the overall architecture of the M2HVideo framework.
The training objectives, including the diffusion model loss and mirror loss, are then detailed (Section~\ref{method:sec3.4}), where the mirror loss, operating in the pixel space, is designed to preserve fine-grained facial details. Finally, we describe the training and inference procedures for our model in Section \ref{method:sec3.5}.

\subsection{Preliminaries}
\label{method:sec3.1}
To make this paper self-contained, we provide a brief overview of latent video diffusion models.
By modeling the reverse process of iterative denoising, latent video diffusion models \cite{blattmann2023stable,liu2024sora,xing2024make} are an efficient framework for high-fidelity video generation.
Unlike standard diffusion models \cite{ho2020denoising,song2020denoising}, which operate directly in high-dimensional pixel spaces, latent video diffusion models leverage a pre-trained variational autoencoder to map high-dimensional $N$ frames of input video $\mathbf{v} \in \mathbb{R}^{N \times H \times W \times C}$ into a compact latent representation $\mathbf{z} \in \mathbb{R}^{N \times h \times w \times c}$, where $h \ll H$ and $w \ll W$. 
Latent video diffusion models have two stages: a forward diffusion process and a reverse denoising process. In the forward process, a sequence of latent codes $\mathbf{z}_1, \cdots, \mathbf{z}_t, \cdots,\mathbf{z}_T$ are
generated by incrementally adding Gaussian noise $\boldsymbol{\epsilon}$ to the initial latent code $\mathbf{z}_0$ which is equal to $\mathbf{z}$ in the forward process. 
This process is defined by:
\begin{equation}
q(\mathbf{z}_t | \mathbf{z}_0) = \mathcal{N}(\mathbf{z}_t; \sqrt{\bar{\alpha}_t} \mathbf{z}_0, (1 - \bar{\alpha}_t) \mathbf{I}),
\end{equation}
where $\bar{\alpha}_t$ denotes the noise schedule and $\mathbf{I}$ is the identity matrix, ensuring isotropic noise addition.
The denoising process aims to learn a mapping function $\boldsymbol{\epsilon}_{\boldsymbol{\theta}}(\mathbf{z}_t, t, \mathbf{c})$ that is trained to predict the added noise, where $\mathbf{c}$ is a conditioning variable guiding the generation of the video.

\subsection{Problem Formulation}
\label{method:sec3.2}
Online retailers commonly use mannequins to present apparel, but such displays lack the realism and emotional resonance of human models. To overcome this limitation, we propose transforming mannequin videos into photorealistic human videos with a specified identity. The synthesized human preserves the original clothing while incorporating identity features from a reference image, enhancing visual appeal and consumer engagement.
Given the physical setup of a mannequin dressed in real clothing, we capture a horizontal surround video encircling the mannequin. This video, denoted by $\mathbf{v}_{m}$, serves as the input along with a target identity image $\mathbf{x}_{id}$. Our objective is to synthesize a new video, $\mathbf{v}_{g}$, in which the identity from $\mathbf{x}_{id}$ is seamlessly integrated into the human regions of $\mathbf{v}_{m}$, while preserving the original dynamics of the clothing and pose.
The core challenge in achieving high-fidelity realism lies in learning a guided mapping that accurately re-renders the characteristics of the identity from $\mathbf{x}_{id}$ onto the non-clothing regions of $\mathbf{v}_{m}$ to ensure visually seamless integration. Formally, this mapping is defined by: $\mathcal{F}: (\mathbf{v}_{m}, \mathbf{x}_{id}) \mapsto \mathbf{v}_{g}$, where $\mathbf{x}_{id} \in \mathbb{R}^{H \times W \times 3}$ is the target identity image, and both $\mathbf{v}_{m}$ and $\mathbf{v}_{g}$ reside in $\mathbb{R}^{N \times H \times W \times 3}$, with $N$ representing the number of video frames.
To facilitate the generation of $\mathbf{v}_{g}$, we employ a pre-trained segmentation model\footnote{\url{https://github.com/Gaoyiminggithub/Graphonomy}} to extract the clothing region $M_c$ from $\mathbf{v}_{m}$, where $M_c \in \mathbb{R}^{N \times H \times W \times 1}$ and each element of $M_c$ takes values in $\{0, 1\}$.
Meanwhile, the body pose $\mathbf{p}_{m}$ from the mannequin video $\mathbf{v}_{m}$ and the facial pose $\mathbf{p}_{id}$, also known as the facial landmarks, from the target identity image $\mathbf{x}_{id}$ are extracted using a pre-trained pose estimation model.\footnote{\url{https://github.com/IDEA-Research/DWPose}}
It should be noted here that $\mathbf{p}_{m}$ is made up of a sequence of $N$ pose images, while $\mathbf{p}_{id}$ corresponds to a single facial pose image.
The learned mapping $\mathcal{F}$ is designed to generate a video $\mathbf{v}_{g}$ that preserves the appearance of the clothing from $\mathbf{v}_{m}$, accurately reflects the facial identity in $\mathbf{x}_{id}$, and ensures visual-temporal fidelity of the generated videos.

\begin{figure*}
\centering
\includegraphics[width=1\textwidth]{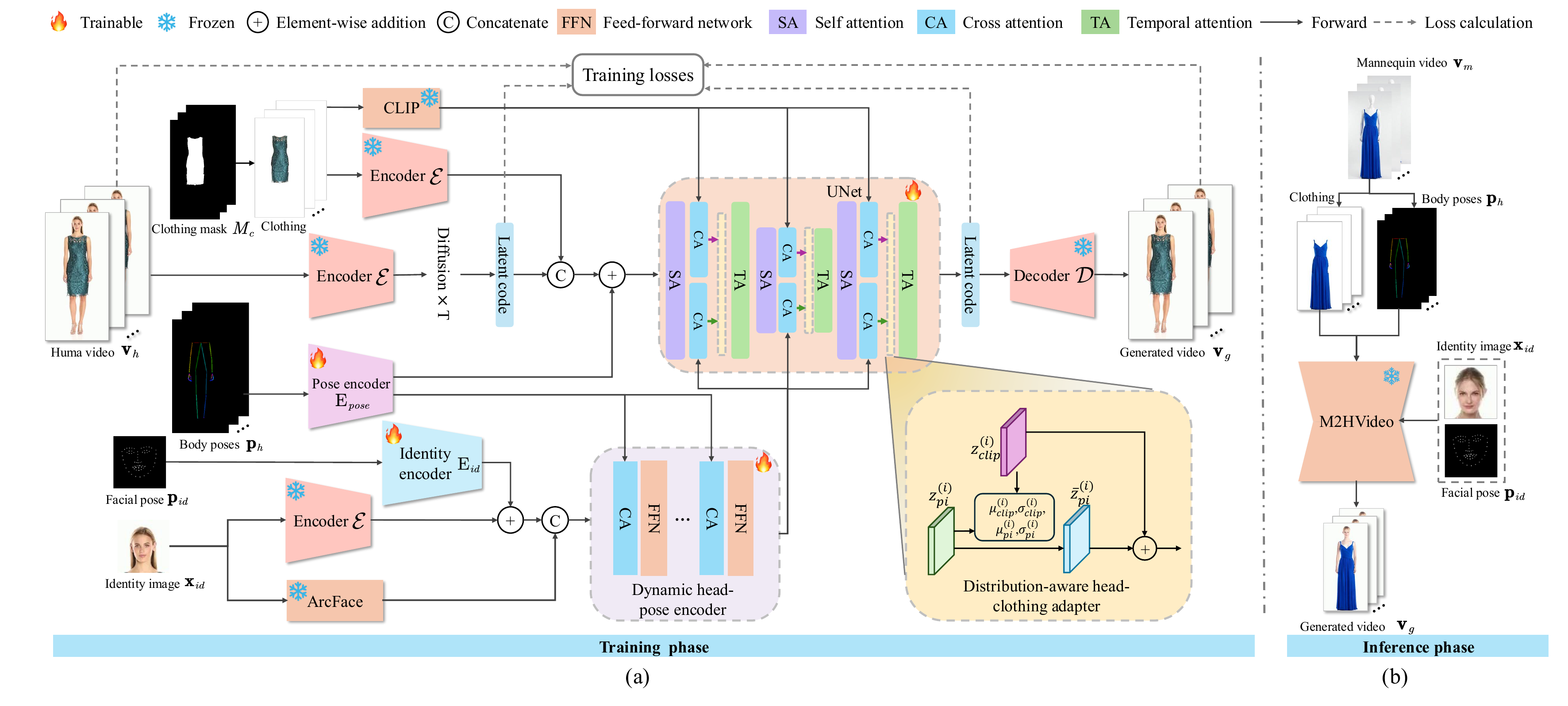}
\caption{Overview of the M2HVideo framework.
(a) Training phase of M2HVideo using human videos.
The body pose, identity image, and pose extracted from a human video are first aligned using the dynamic pose-aware head encoder. These features, along with clothing representations, are passed to a UNet to synthesize photorealistic human videos. The output is supervised by the loss function in both the latent and pixel spaces to ensure consistency and realism of the appearance.
(b) Inference phase of M2HVideo with the mannequin video and identity image as input. 
Each frame of the mannequin video is processed to extract the clothing and body pose, which are then combined with an identity image and fed into our model to produce realistic human videos conditioned on the mannequin's clothing and the target identity.}
\label{md_fig:1}
\end{figure*}

\subsection{M2HVideo Framework}
\label{method:sec3.3}
\textbf{Overview of the Framework.}
The proposed M2HVideo framework aims to generate a photorealistic human video $\mathbf{v}_{g}$ by taking a mannequin video $\mathbf{v}_{m}$ and a target identity image $\mathbf{x}_{id}$ as inputs, in such a way that $\mathbf{v}_{g}$ preserves the clothing from $\mathbf{v}_{m}$ while reflecting the identity represented in $\mathbf{x}_{id}$.
Due to the lack of triplet data $\{\mathbf{v}_{m}, \mathbf{x}_{id}, \mathbf{v}_{g}\}$, we train the framework by human videos $\mathbf{v}_{h}$, treating their clothing masks $M_c$ and body poses $\mathbf{p}_{h}$ as proxies for the attributes of a mannequin. An overview of the proposed framework is shown in Fig. \ref{md_fig:1}.
During the training phase, as illustrated in Fig. \ref{md_fig:1}(a), the input to the framework consists of two components: the latent code fed into the UNet, and the conditioning signals $\mathbf{c}$ provided to guide the UNet.
To obtain the latent code, each frame of $\mathbf{v}_{h}$ is first encoded into a latent space using a frozen VAE encoder $\mathcal{E}$, and Gaussian noise is added to obtain the diffusion latent code $\mathbf{z}_{t}$, which serves as the primary input to the UNet denoising model. To inject the appearance of the clothing, the binary mask $M_c$ is multiplied element-wise by $\mathbf{v}_{h}$ to isolate the clothing region, which is then encoded by $\mathcal{E}$ to yield the clothing embedding $\mathbf{f}_{c}$; this is concatenated with the diffusion latent code to form a unified appearance representation. Information about the body pose is incorporated by encoding $\mathbf{p}_{h}$ using a pose encoder $E_{pose}$, whose output is added element-wise to the unified appearance representation to produce the final latent input to the UNet. 
In addition, so as to incorporate conditioning signals into the UNet, the reference image $\mathbf{x}_{id}$ is processed by the frozen VAE encoder $\mathcal{E}$ to extract the identity's latent code, which is temporally duplicated across frames. An identity encoder $E_{id}$ also encodes the facial pose $\mathbf{p}_{id}$, producing a facial pose embedding that is likewise duplicated and added element-wise to the identity's latent code. These features are concatenated with a facial embedding extracted from $\mathbf{x}_{id}$ via ArcFace to form the final identity representation. This identity representation is fused with the body pose embedding through a dynamic pose-aware head encoder $\mathcal{H}$, producing a pose-aware identity embedding $\mathbf{f}_{pi}$. 
Along with a CLIP-based clothing embedding, $\mathbf{f}_{pi}$ is fed into a distribution-aware head-clothing adapter $\mathcal{A}$ to address shifts in the distribution and enhance the temporal consistency.
The denoised latent codes are finally decoded by the VAE decoder $\mathcal{D}$ into videos.
During inference, as shown in Fig. \ref{md_fig:1}(b), the clothing mask $M_c$ and body pose $\mathbf{p}_{m}$ are extracted from $\mathbf{v}_{m}$ and used with $\mathbf{x}_{id}$ as input to the generation module $\mathcal{F}$ to synthesize the final video $\mathbf{v}_{g}$. Details on the modules $\mathcal{H}$ and $\mathcal{A}$ are given below.

\textbf{Dynamic Pose-Aware Head Encoder.}
In our task, the sole human-related input is an identity image $\mathbf{x}_{id}$, making ambiguous the precise localization of the head in the generated videos.
However, the body pose $\mathbf{p}_h$ offers spatial cues for estimating the head's position.
To exploit this, we introduce a dynamic pose-aware head encoder that guides the denoising UNet in accurately determining the placement of both the identity and the head.
Specifically, the identity image $\mathbf{x}_{id}$ and its facial pose $\mathbf{p}_{id}$, along with the body pose $\mathbf{p}_{h}$, are input into the encoder to learn the dependency the position of the head on the body pose, enhancing the accuracy of the localization.
As shown in Fig. \ref{md_fig:1}(a), $\mathbf{x}_{id}$ is encoded into a latent identity code via a frozen encoder $\mathcal{E}$, while $\mathbf{p}_{id}$ and $\mathbf{p}_{h}$ are embedded using $E_{id}$ and $E_{pose}$. 
The identity code and facial pose embedding are summed and temporally duplicated to match the frame count of $\mathbf{p}_{h}$.
Additionally, a facial embedding from ArcFace is extracted from $\mathbf{x}_{id}$ to emphasize identity-specific features.
The duplicated feature and ArcFace embeddings are concatenated and combined with the body pose embedding via a four-layer encoder having cross-attention and feed-forward modules. 
Notably, the concatenated embeddings serve as queries, while the body pose embedding functions as both key and value in the cross-attention layers. The output of this encoder is a pose-aware identity embedding $\textbf{f}_{pi}$, which is subsequently used to condition the denoising UNet.
The output $\textbf{f}_{pi}$ of the dynamic pose-aware head encoder $\mathcal{H}$ is formally defined by:
\begin{equation}
\textbf{f}_{pi} = \mathcal{H}\left(\operatorname{concat}\left(\mathcal{E}(\mathbf{x}_{id}) + E_{id}(\mathbf{p}_{id}), \text{Arc}(\mathbf{x}_{id})\right),E_{pose}(\mathbf{p}_h)\right),
\label{eq.pose_aware}
\end{equation}
where $\operatorname{concat}(\cdot, \cdot)$ denotes the concatenate operation in the channel dimension. The pose-aware identity embedding $\textbf{f}_{pi}$ is fed into UNet along with the clothing features to improve the consistency between the appearance of the head and that of the clothing, which is further refined by our proposed distribution-aware head-clothing adapter.

\textbf{Distribution-Aware Head-Clothing Adapter.}
In the denoising UNet architecture, temporal layers placed after the spatial layers disrupt the spatial distribution of the latent representations, which leads to distortions of the features and visual artifacts such as blurring and deformation.
To resolve this problem, we propose a distribution-aware head-clothing adapter to align the distributions of the features of the identity and clothing before each temporal layer. This preserves the spatial consistency and reduces the occurrence of artifacts.
We modify the spatial attention blocks in the UNet architecture. For the output $\mathbf{z}_i$ of the self-attention module, we compute the spatial cross-attention with both the pose-aware identity embedding $\mathbf{f}_{pi}$ and the clothing embedding $\mathbf{f}_{clip}$, which is extracted from the clothing sequence by CLIP. 
This results in intermediate features $\mathbf{z}^{(i)}_{pi}$ and $\mathbf{z}^{(i)}_{clip}$ in the $i$-th spatial attention blocks.
Due to the distributional discrepancy between $\mathbf{z}_{pi}^{(i)}$ and $\mathbf{z}_{clip}^{(i)}$, directly adding them can lead to misalignment. To mitigate this, we normalize $\mathbf{z}_{pi}^{(i)}$ to match the distribution of $\mathbf{z}_{clip}^{(i)}$ using an affine transformation:
\begin{equation}
\bar{\mathbf{z}}_{pi}^{(i)} = \frac{\mathbf{z}_{pi}^{(i)} - \mu_{pi}^{(i)}}{\sigma_{pi}^{(i)}} \cdot \sigma_{clip}^{(i)} + \mu_{clip}^{(i)},
\end{equation}
where $\mu$ and $\sigma$ denote the mean and standard deviation of the corresponding feature embeddings. This distributional alignment ensures seamless fusion between $\bar{\mathbf{z}}_{pi}^{(i)}$ and $\mathbf{z}_{clip}^{(i)}$, reducing integration artifacts and improving the visual consistency of the generated frames.

\subsection{Training Losses}
\label{method:sec3.4}
Given a clothing mask $M_c$, a sequence of body poses $\mathbf{p}_{h}$ from $\mathbf{v}_{h}$, an
identity image $\mathbf{x}_{id}$ which is randomly selected from the frames in $\mathbf{v}_{h}$, and its facial pose $\mathbf{p}_{id}$, we train our framework using the following
loss functions: 
$\mathcal{L}_{\text{diff}}$ for latent noise prediction error, and $\mathcal{L}_{\text{mir}}$ for image-level consistency to alleviate latent space recovery limitations of post-compression.

\textbf{Diffusion Loss.}
Our primary objective is to ensure that the distribution of the generated video aligns closely with that of the human video $\mathbf{v}_h$ in the latent space. To achieve this, we employ the latent diffusion loss $\mathcal{L}_{\text{diff}}$, which is widely used in generative frameworks based on diffusion models. The conditional input $\mathbf{c}$ to the denoising network includes the clothing mask, the identity image, the pose information, and the pose prior, denoted by $\{M_c, \mathbf{x}_{id}, \mathbf{p}_{id}, \mathbf{p}_{h}\}$. Therefore, the latent diffusion loss $\mathcal{L}_{\text{diff}}$ can be formalized as:
\begin{equation}
\mathcal{L}_{\text {diff }}=\mathbb{E}_{\mathbf{z}_0, \boldsymbol{\epsilon}, t}\left\|\boldsymbol{\epsilon}-\boldsymbol{\epsilon}_{\boldsymbol{\theta}}\left(\mathbf{z}_t, t,\left\{M_c, \mathbf{x}_{i d}, \mathbf{p}_{i d}, \mathbf{p}_h\right\}\right)\right\|_2^2.
\label{eq.diff}
\end{equation}
This loss encourages the denoising network to accurately predict the added noise at each diffusion timestep $t$, conditioned on the semantic and structural guidance provided by the input set
$\{M_c, \mathbf{x}_{id}, \mathbf{p}_{id}, \mathbf{p}_{h}\}$.

\textbf{Mirror Loss.}
During the diffusion stage, the input videos are first mapped into the latent space via the encoder of a pre-trained VAE, and subsequently reconstructed in the pixel space using the corresponding VAE decoder. However, this encode–decode process introduces non-trivial degradation, particularly in semantically rich yet spatially small regions, such as human faces. To mitigate this, we propose a hybrid latent-to-pixel strategy designed to restore fine-grained facial details.
To further enhance the fidelity in high-frequency regions, we introduce the mirror loss in the pixel space. However, computing the pixel-space loss directly on the final denoised output $\mathbf{z}_0$ is computationally expensive during training, as it requires multiple iterations of reverse denoising. We approximate the final denoised latent $\tilde{\mathbf{z}}_0$ using a one-step denoising formula:
\begin{equation}
\label{eq.denois}
\tilde{\mathbf{z}}_0 = \frac{\mathbf{z}_t}{\sigma^2 + 1} - \frac{\sigma}{\sqrt{\sigma^2 + 1}} \boldsymbol{\epsilon}_{\boldsymbol{\theta}},
\end{equation}
where $\sigma$ denotes the standard deviation of the noise sampled via a cosine schedule, and $\mathbf{\epsilon}_{\boldsymbol{\theta}}$ is the predicted noise. This approximation enables efficient supervision while maintaining scalability. The approximated latent $\tilde{\mathbf{z}}_0$ is then decoded through $\mathcal{D}$ to obtain the generated video $\mathbf{v}_{g}$.
To preserve fine-grained details, particularly in facial regions, we introduce a mirror loss that constrains the generated results in both the pixel and perceptual dimensions. In particular, we adopt an L2 loss in the pixel space to enforce direct correspondence between the generated and the ground-truth pixels, and a perceptual loss computed based on a pre-trained VGG network \cite{simonyan2014very}, denoted as $\phi$,
to encourage semantic similarity. 
Formally, the mirror loss is defined by:
\begin{equation}
\label{eq.mirror}
\begin{aligned}
\mathcal{L}_{\text {mir}}= & \alpha \sum_{i=1}^N\left\|\left(\mathbf{v}_h^{(i)}-\mathbf{v}_g^{(i)}\right) \odot M_{i d}^{(i)}\right\|_2^2 +\\
& \beta \sum_{i=1}^N \sum_l\left\|\phi_l\left(\mathbf{v}_h^{(i)} \odot M_{i d}^{(i)}\right)-\phi_l\left(\mathbf{v}_g^{(i)} \odot M_{i d}^{(i)}\right)\right\|_2^2,
\end{aligned}
\end{equation}
where $\mathbf{v}_{h}^{(i)}$ and $\mathbf{v}_{g}^{(i)}$ denote the $i$-th frames of the ground-truth and generated videos, respectively. $M_{id}^{(i)}$ denotes the $i$-th frame of the facial region mask obtained using ArcFace, $l \in \{relu1\_2, {relu2\_2}, {relu3\_4}\}$ is the layer of VGG-16, $\phi_l(\cdot)$ denotes the function of layer $l$, and `$\odot$' denotes element-wise multiplication. 
The coefficients $\alpha$ and $\beta$ control the contributions of the pixel-level reconstruction loss and perceptual loss.

\textbf{Total Loss.}
For training our M2HVideo, the full loss function can be summarized as follows:
\begin{equation}
\mathcal{L}_{\text{total}} = \mathcal{L}_{\text{diff}} + \mathcal{L}_{\text{mir}},
\label{eq.total}
\end{equation}
where $\mathcal{L}_{\text{diff}}$ denotes the latent diffusion loss, which directs the denoising network to estimate the noise introduced at each diffusion timestep accurately, and the term $\mathcal{L}_{\text{mir}}$ focuses on facial regions to improve the fidelity of the details and to maintain the consistency of the identity.

\begin{algorithm}[t]
\caption{
Training algorithm for the proposed M2HVideo framework.
}
\label{train_alg}
\small
\SetAlgoLined
\KwIn{Human video dataset $\{(\mathbf{v}_{h}, \mathbf{p}_{h}, \mathbf{x}_{id}, \mathbf{p}_{id}, M_c, M_{id})\}$.}
\KwOut{
Optimized generation module $\mathcal{F}$ of M2HVideo.
}
\linespread{1.2}\selectfont 
\vspace{0.5em}
Initialize the parameters $\boldsymbol{\theta}_{E_{id}}, \boldsymbol{\theta}_{E_{pose}}, \boldsymbol{\theta}_{\mathcal{H}}, \boldsymbol{\theta}_{\mathcal{A}}$ corresponding to the modules $E_{id}$, $E_{pose}$, $\mathcal{H}$, $\mathcal{A}$, respectively; \\
Load the parameters $\boldsymbol{\theta}_{\text{UNet}}$ of the UNet;\\
Load the parameters of $\mathcal{E}$, $\mathcal{D}$, $\text{Arc}$, $\text{CLIP}$, and $\phi$ from pre-trained models;\\
\vspace{0.5em}
\For{$iter\leftarrow 1$ \KwTo $N_{iter}$}{
Sample a batch of data: $\mathbf{v}_{h}$, $\mathbf{p}_{h}$, $\mathbf{x}_{id}$, $\mathbf{p}_{id}$, $M_c$, and $M_{id}$;\\

\If{$iter \leq \frac{N_{iter}}{2}$}{
\textcolor[rgb]{0.5,0.5,0.5}{\# Downsample by a factor of 0.5}\\
$\mathbf{v}_{h} \leftarrow \operatorname{Downsample}(\mathbf{v}_{h}, 0.5)$; \textcolor[rgb]{0.5,0.5,0.5}{\# The spatial dimensions (width and height) of $M_c$, $\mathbf{p}_{h}$, $\mathbf{x}_{id}$, and $\mathbf{p}_{id}$ are adjusted accordingly}\\
}

$\boldsymbol{\epsilon} \sim \mathcal{N}(0, \mathbf{I})$, $t \sim \text{Uniform}(1, T)$;\\
$\mathbf{z}_t=\sqrt{\bar{\alpha}_{t}}\mathcal{E}(\mathbf{v}_{h})+\sqrt{1-\bar{\alpha}_{t}}\boldsymbol{\epsilon};$ 
\textcolor[rgb]{0.5,0.5,0.5}{\# Forward process of diffusion model, more details can be found in  \cite{ho2020denoising}}

$\mathbf{f}_{c}=\mathcal{E}(\mathbf{v}_{h} \odot M_c)$;

$\mathbf{f}_{clip}=\text{CLIP}(\mathbf{v}_{h} \odot M_c)$;

$\textbf{f}_{pi} = \mathcal{H}\left(\operatorname{concat}\left(\mathcal{E}(\mathbf{x}_{id}) + E_{id}(\mathbf{p}_{id}), \text{Arc}(\mathbf{x}_{id})\right),E_{pose}(\mathbf{p}_h)\right);$

$\boldsymbol{\epsilon}_{\boldsymbol{\theta}}=\text{UNet}(\operatorname{concat}(\mathbf{z}_{\text{t}},\mathbf{f}_{c})+E_{pose}(\mathbf{p}_{h}),\mathcal{A}(\mathbf{f}_{pi},\mathbf{f}_{clip}))$;

$\tilde{\mathbf{z}}_0 = \frac{\mathbf{z}_t}{\sigma^2 + 1} - \frac{\sigma}{\sqrt{\sigma^2 + 1}} \boldsymbol{\epsilon}_{\boldsymbol{\theta}}$;
\textcolor[rgb]{0.5,0.5,0.5}{\# See Eq. (\ref{eq.denois})}\\

$\mathbf{v}_{g}=\mathcal{D}(\tilde{\mathbf{z}}_0)$\;

$\mathcal{L}_{\text{diff}} = \mathbb{E}_{\mathbf{z}_0, \boldsymbol{\epsilon}, t}  \left\lVert \boldsymbol{\epsilon} - \boldsymbol{\epsilon}_{\boldsymbol{\theta}}(\mathbf{z}_t, t, \{M_c, \mathbf{x}_{id}, \mathbf{p}_{id}, \mathbf{p}_{h}\}) \right\rVert^2_2 ;$ 

$\mathcal{L}_{\text {mir}}=  \alpha \sum_{i=1}^N\left\|\left(\mathbf{v}_h^{(i)}-\mathbf{v}_g^{(i)}\right) \odot M_{i d}^{(i)}\right\|_2^2 +
 \beta \sum_{i=1}^N \sum_l\left\|\phi_l\left(\mathbf{v}_h^{(i)} \odot M_{i d}^{(i)}\right)-\phi_l\left(\mathbf{v}_g^{(i)} \odot M_{id}^{(i)}\right)\right\|_2^2;$

$\mathcal{L}_{\text{total}} = \mathcal{L}_{\text{diff}} + \mathcal{L}_{\text{mir}} $;
\textcolor[rgb]{0.5,0.5,0.5}{\# See Eq. (\ref{eq.total})}\\

$\boldsymbol{\theta}_\text{learnable} \leftarrow \boldsymbol{\theta}_\text{learnable} - \eta \nabla_{\boldsymbol{\theta}_\text{learnable}}\mathcal{L}_{\text{total}};$
\textcolor[rgb]{0.5,0.5,0.5}{\# Update the learnable parameters, i.e., $\boldsymbol{\theta}_{E_{id}}, \boldsymbol{\theta}_{E_{pose}}, \boldsymbol{\theta}_{\mathcal{H}}, \boldsymbol{\theta}_{\mathcal{A}}, \boldsymbol{\theta}_{\text{UNet}}$}\\
}
\Return{$\mathcal{F}^{\ast}$}; \textcolor[rgb]{0.5,0.5,0.5}{\# Optimized $\mathcal{F}$}

\end{algorithm}

\subsection{The Training Process}
\label{method:sec3.5}
This subsection details the training process designed to optimize our proposed framework. The complete training pipeline is outlined in \textbf{Algorithm} \ref{train_alg}. At first, the parameters of the identity encoder $E_{id}$, the pose encoder $E_{pose}$, the dynamic pose-aware head encoder $\mathcal{H}$, and the distribution-aware head-clothing adapter $\mathcal{A}$ are initialized, while the parameters of the UNet, $\mathcal{E}$, $\mathcal{D}$, and Arc, CLIP, and $\phi$ are loaded from the pre-trained models (shown in lines 1-3). 
To enhance the efficiency of the training, we adopt a progressive resolution strategy. 
In particular, the model is trained on half-resolution videos (shown in lines 6–9) during the first half of the $N_{iter}$ iterations, and subsequently on the full-resolution videos.
In each training iteration, we employ the forward process of the denoising diffusion probabilistic model (DDPM) \cite{ho2020denoising} to add noise to the latent codes. In particular, noise $\boldsymbol{\epsilon}$ is sampled from a standard normal distribution $\mathcal{N}(0,\mathbf{I})$, and the timestep $t$ is drawn from a uniform distribution.
Using the noise schedule parameter $\bar{\boldsymbol{\alpha}}_t$, a noisy latent code $\mathbf{z}_t$ is generated (shown in lines 10-11). 
Clothing-related features $\mathbf{f}_{c}$ and $\mathbf{f}_{clip}$ are extracted by encoding the masked clothing video, formed by the element-wise multiplication of the human video $\mathbf{v}_h$ and the clothing mask $M_c$, by $\mathcal{E}$ and CLIP, respectively (shown in lines 12-13). Pose features $E_{pose}(\mathbf{p}_h)$ and identity features $\mathcal{E}(\mathbf{x}_{id})$, $E_{pose}(\mathbf{p}_{id})$, and Arc$(\mathbf{p}_{id})$ are extracted and fused by the dynamic pose-aware head encoder $\mathcal{H}$ to produce a pose-aware identity embedding $\mathbf{f}_{pi}$ for conditioning the generation (shown in line 14).
The embeddings $\mathbf{f}_{pi}$ and $\mathbf{f}_{clip}$ are integrated into the UNet via the distribution-aware head–clothing adapter $\mathcal{A}$, which mitigates distributional discrepancies and promotes temporal coherence. A one-step denoising strategy is then applied to obtain the denoised latent code $\tilde{\mathbf{z}}_0$, which is subsequently decoded by $\mathcal{D}$ to reconstruct the generated video $\mathbf{v}_g$ in the pixel space (shown in lines 16-17). 
Then, our framework is trained with the combined loss $\mathcal{L}_{\text{diff}}$ and $\mathcal{L}_{\text{mir}}$ (shown in lines 18–21). When the training process converges, our method returns the optimized generation module $\mathcal{F}$, which synthesizes videos that preserve the clothing in $\mathbf{v}_h$ and reflect the identity in $\mathbf{x}_{id}$.

\section{Experiments}
\label{experiment}
In this section, we first introduce the construction of the datasets used in this research and outline the details of the implementation. Then, we experimentally demonstrate the superiority of our proposed framework over existing methods. Furthermore, we conduct a comprehensive ablation study to evaluate the individual contributions of the key components within our framework.

\subsection{Datasets}
\label{exp:dataset}
\begin{figure}[t]
    \centering
    \includegraphics[width=0.48\textwidth]{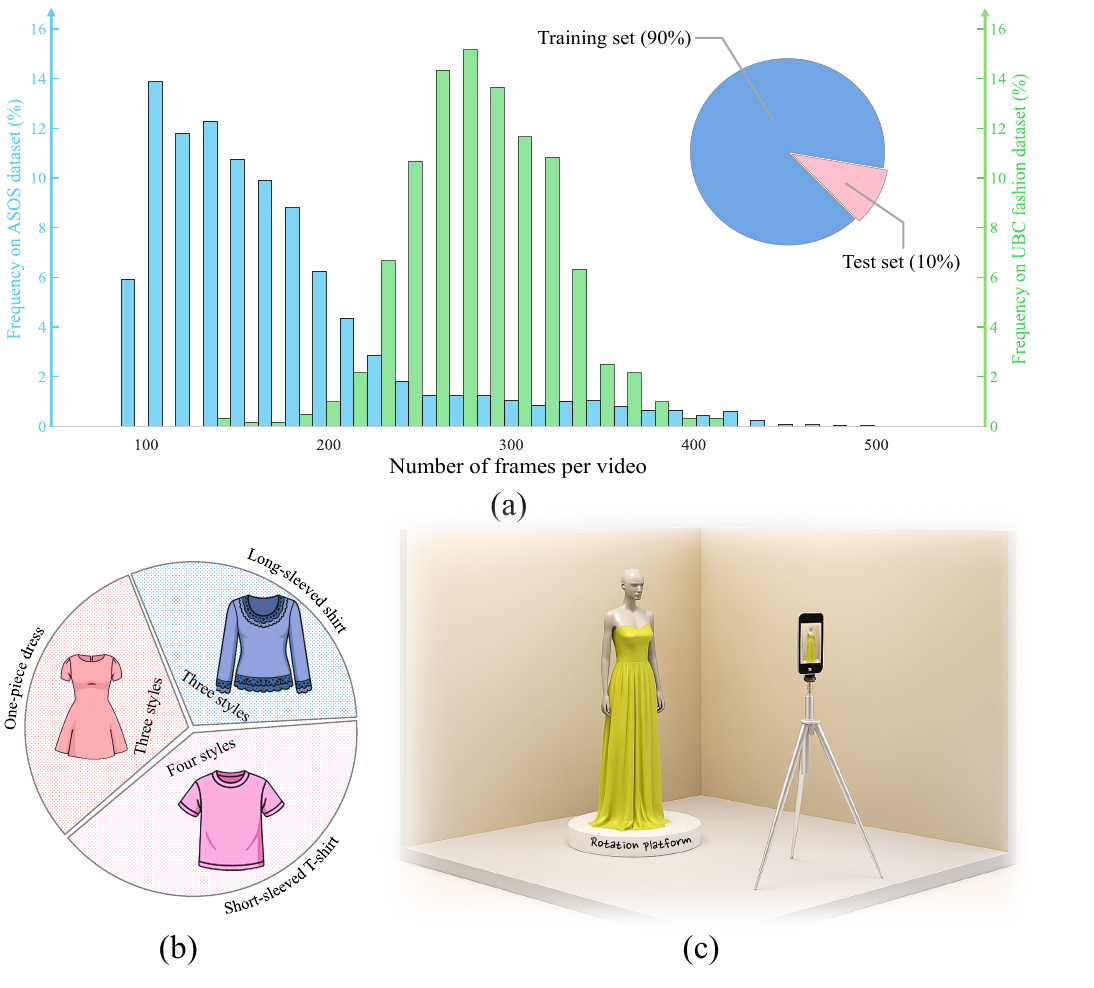}
    \caption{
    Statistics and visualizations of the datasets used in this study. (a) Distribution of video frame counts and data partitioning for the UBC fashion and ASOS datasets. (b) Clothing category distribution in the MannequinVideos dataset, illustrating the proportions of T-shirts, long-sleeved shirts, and dresses. (c) Data acquisition setup for the MannequinVideos dataset, showing the studio environment with a mannequin on a rotating platform and a smartphone used for capturing the video.
    }
    \label{fig:dataset_constr}
\end{figure}

To evaluate our proposed framework for the mannequin-to-human video generation task, we use two real-model datasets and one mannequin dataset: the UBC fashion dataset \cite{zablotskaia2019dwnet}, the ASOS dataset, and a newly collected mannequin dataset called MannequinVideos. As illustrated in Fig. \ref{fig:dataset_constr}(a), the UBC fashion and ASOS datasets exhibit distinct distributions in video frame counts and are partitioned into training and test splits to facilitate comprehensive evaluation. 
In particular, the UBC fashion dataset consists of 600 studio-captured videos, each with a resolution of $940\times720$. All videos were recorded by a fixed camera against a pure white background, featuring models who perform slight lateral movements followed by a full turn. 
In comparison to the UBC fashion dataset, the ASOS dataset features more dynamic and diverse movements of the models, including walking toward the camera and turning from side to side. Both datasets are randomly split into 90\% for training and 10\% for testing. 
To further assess the effectiveness of our method on real-world mannequins, we constructed a new dataset, MannequinVideos. The dataset is made up of videos of mannequins in four distinct poses, each dressed in one of the three following types of clothing: T-shirts, long-sleeved shirts, and dresses. Figure \ref{fig:dataset_constr}(b) shows the clothing distribution, including four distinct styles of T-shirts, three styles of long-sleeved shirts, and three styles of one-piece dresses. To maintain the consistency of the outfits, both short-sleeved T-shirts and long-sleeved shirts were randomly paired with shorts or long pants.
To collect the data, we built a studio and placed the mannequin at the center of a rotating platform to record a multi-view video, as depicted in Fig. \ref{fig:dataset_constr}(c). For instance, the mannequin rotates from 70 degrees to the right to 70 degrees to the left, providing a wide range of perspectives. All videos were recorded in-studio at a resolution of $512 \times 512$, with each video consisting of 60 frames.

\subsection{Implementation Details}
Since the three datasets have different resolutions, we retained the original video lengths and applied center cropping along the width to enforce a $2{:}1$ aspect ratio for both the UBC fashion dataset and MannequinVideos. The cropped videos were then resized to a resolution of $512 \times 256$ to standardize the dimensions of the input.
For each training instance, we sampled eight frames per video by randomly selecting a starting frame and subsequently sampling one frame every four frames.
During the inference, we employed the DDIM sampler with 50 steps and used a classifier-free guidance, applying a scale factor of 7.5 to synthesize photorealistic human videos. 
The weighting coefficients $\alpha$ and $\beta$ in Eq. (\ref{eq.mirror}) were set to 0.05 and 0.001, respectively.
The training of the model was conducted on two NVIDIA A6000 GPUs, with a batch size of four per GPU. The training process had 27,500 iterations on the UBC fashion dataset and 311,750 iterations on the ASOS dataset.
The optimization used the Adam \cite{kingma2014adam} optimizer, with parameters $\beta_1 = 0$, $\beta_2 = 0.99$, and a learning rate $\eta$ of $5 \times 10^{-5}$.

\subsection{Comparisons with State-of-The-Art}
\label{exp:2}

\begin{figure*}[t]
\includegraphics[width=0.96\textwidth]{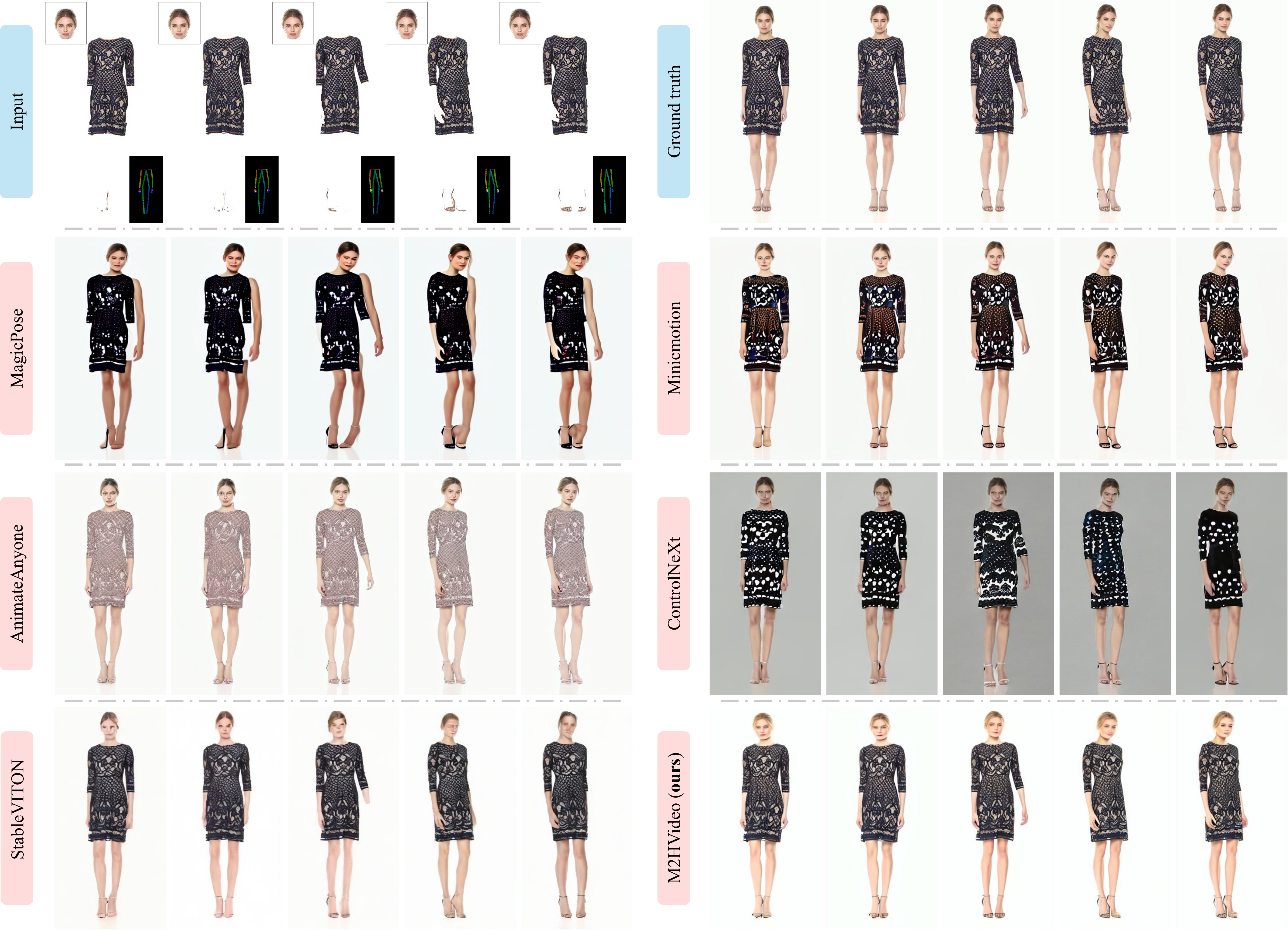}
\caption{Comparisons between our proposed M2HVideo and some baselines, namely, MagicPose \cite{chang2023magicpose}, MimicMotion \cite{zhang2024mimicmotion}, AnimateAnyone \cite{hu2024animate}, StableVITON \cite{kim2024stableviton}, and ControlNeXt \cite{peng2024ControlNeXt} on the UBC fashion dataset.}
\label{fig:UBC}
\end{figure*}

\subsubsection{Baselines}
As mannequin-to-human video generation is a novel task, there are no existing methods that directly address the same objective. Therefore, we compared our proposed framework with five state-of-the-art methods that share similar goals.
The selected baselines include three pose transfer methods: MagicPose \cite{chang2023magicpose}, MimicMotion \cite{zhang2024mimicmotion}, and AnimateAnyone \cite{hu2024animate}; one virtual try-on method, StableVITON \cite{kim2024stableviton}; and one controllable video generation model, ControlNeXt \cite{peng2024ControlNeXt}. To make this paper self-contained, a brief introduction to these baselines follows:

\textbf{MagicPose} is a pose transfer framework based on the ControlNet \cite{zhang2023adding} architecture. It enhances appearance control by integrating the appearance latent feature map with the self-attention layers of the UNet.

\textbf{MimicMotion} is based on the insight that the confidence of pose landmarks is correlated with the visual quality of the generated videos. It dynamically adjusts the loss weights according to the head region's pose confidence to enhance the visual quality.

\textbf{AnimateAnyone} employs a dual-UNet architecture for pose transfer. It extracts appearance features using a reference encoder and uses a lightweight pose guider to align the pose and the appearance.

\textbf{StableVITON} is the first LDM-based virtual try-on framework. It leverages a learnable pre-trained stable diffusion encoder and an attention-guided total variation loss to retain fine clothing details.

\textbf{ControlNeXt} is a lightweight conditional video generation framework. It encodes the control signal using a convolutional module and injects this signal into the denoising UNet via cross-normalization.

\begin{figure*}[t]
\includegraphics[width=0.96\textwidth]{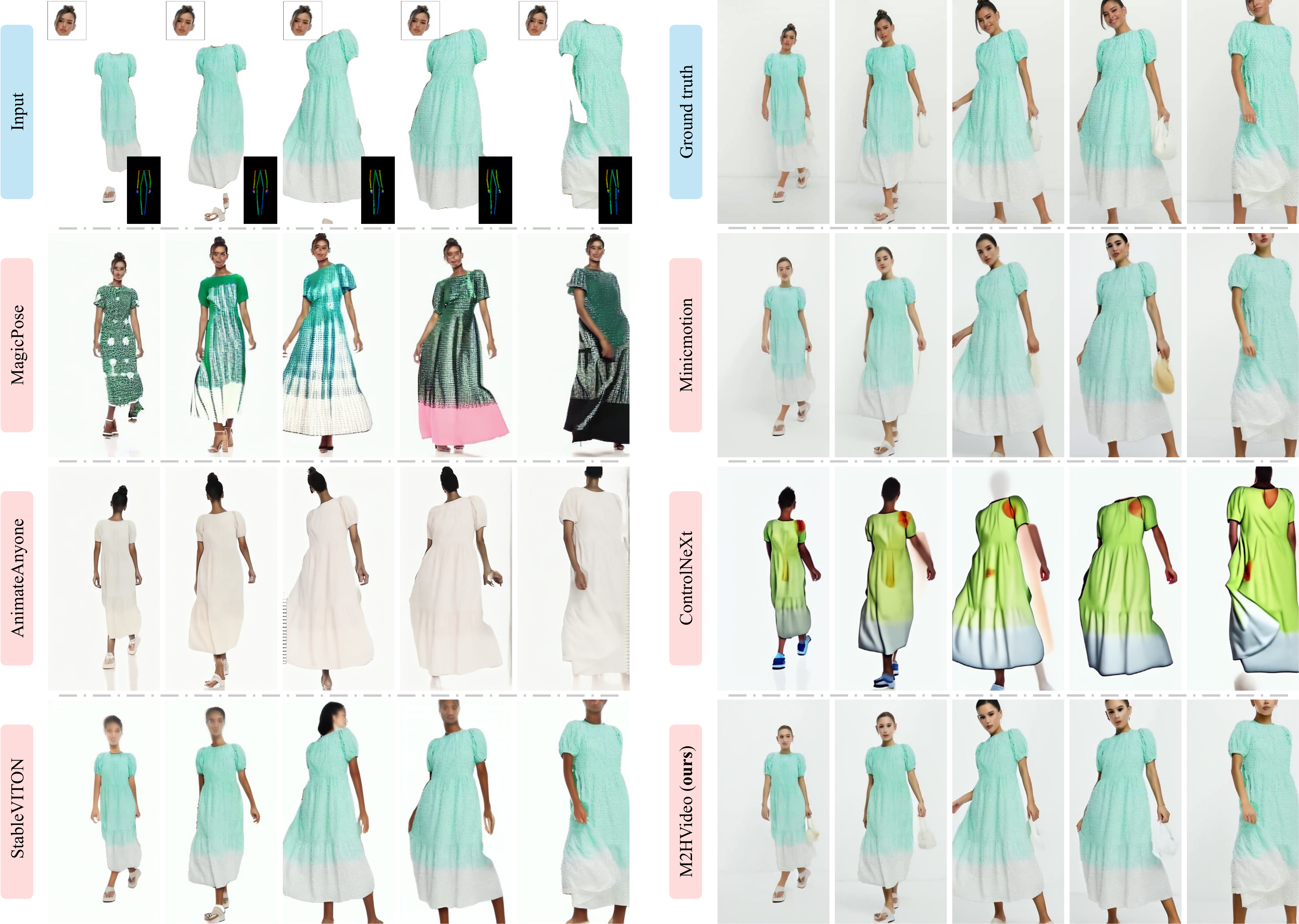}
\caption{Comparisons between our proposed M2HVideo and the baselines, namely, MagicPose \cite{chang2023magicpose}, MimicMotion \cite{zhang2024mimicmotion}, AnimateAnyone \cite{hu2024animate}, StableVITON \cite{kim2024stableviton}, and ControlNeXt \cite{peng2024ControlNeXt} on the ASOS dataset.}
\label{fig:exp_ASOS}
\end{figure*}

\begin{figure}[!t]
\centering
\includegraphics[width=0.48\textwidth]{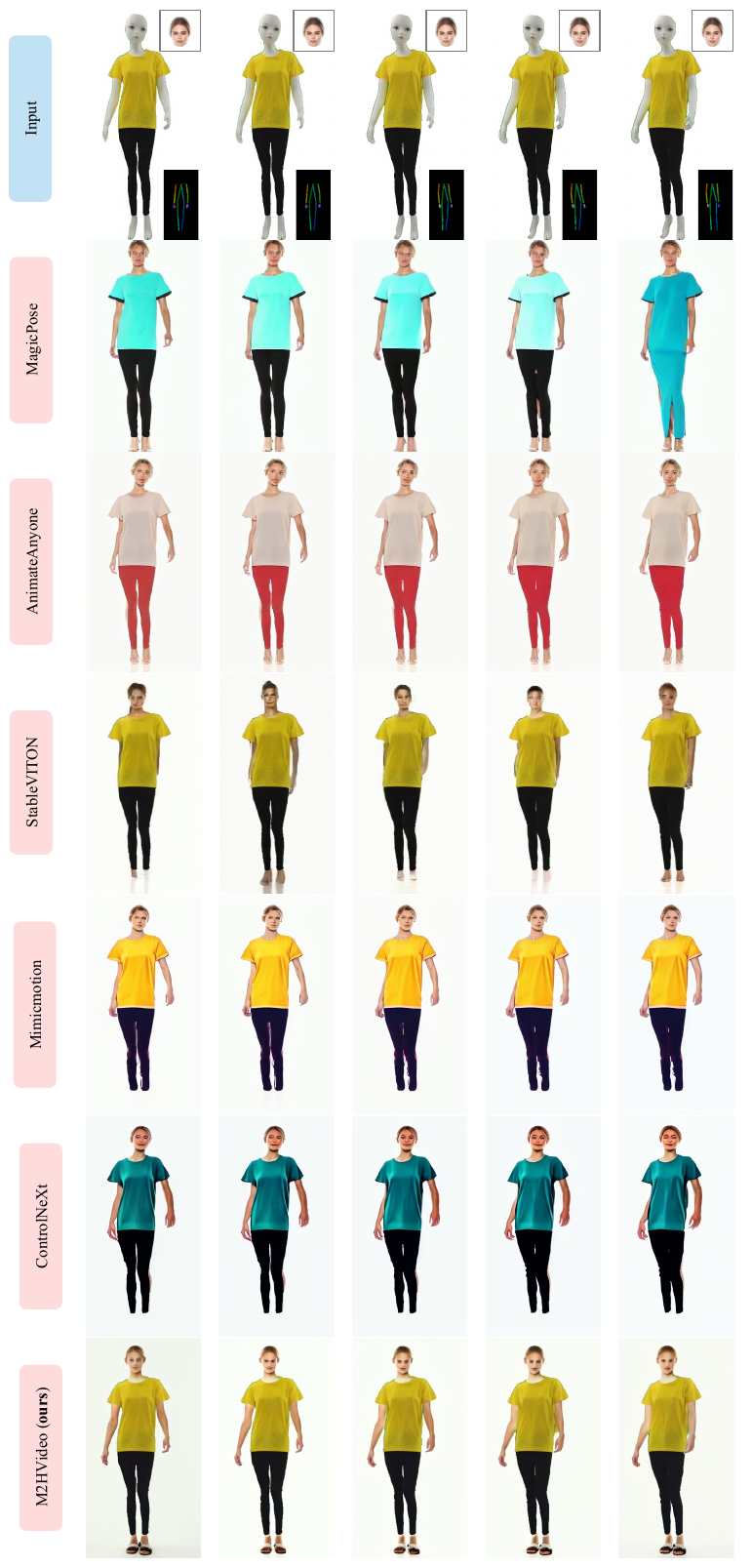}
\caption{Qualitative comparison of M2HVideo with the baselines on the MannequinVideos dataset, demonstrating the translation from mannequin to real human.}
\label{fig:exp_mann}
\end{figure}

\begin{table*}[t]
\caption{Quantitative comparison of M2HVideo and the baselines on the UBC fashion dataset and ASOS dataset in terms of clothing consistency (PSNR, SSIM, and LPIPS), identity preservation (CSIM), and visual-temporal fidelity (FVD); the best is \textbf{bolded} and the second best is \underline{underlined}}
\label{Quantitative_UBC_ASOS_fashion}
\centering
\resizebox{0.95\textwidth}{!}{
\begin{tabular}{lccccccccccc}
\toprule
\multirow{2}{*}{Method} & \multicolumn{5}{c}{UBC fashion dataset} & \multicolumn{5}{c}{ASOS dataset} \\
\cmidrule(lr){2-6} \cmidrule(lr){7-11}
& PSNR ($\uparrow$) & SSIM ($\uparrow$) & LPIPS ($\downarrow$) & CSIM ($\uparrow$) & FVD ($\downarrow$) & PSNR ($\uparrow$) & SSIM ($\uparrow$) & LPIPS ($\downarrow$) & CSIM ($\uparrow$) & FVD ($\downarrow$) \\
\midrule

MagicPose \cite{chang2023magicpose} & 15.22  & 0.656 & 0.237 & 0.913 & 112.75 & 14.89 & 0.63 & 0.213 & 0.908 & 175.36 \\
MimicMotion \cite{zhang2024mimicmotion}  & 23.01 & \underline{0.878} & \underline{0.079} & 0.928 & 35.52 & 23.76 & 0.856 & \underline{0.053} & \underline{0.927} & 38.71 \\ 
AnimateAnyone \cite{hu2024animate}     & 21.72 & 0.858 & 0.103 & \underline{0.932} & 53.93  & 21.35 & 0.781 & 0.078 & 0.902 & 97.64 \\
StableVITON \cite{kim2024stableviton}  & \underline{23.25} & 0.864 & 0.082 & 0.912 & \underline{31.06} & \underline{23.98} & \underline{0.862} & 0.056 & 0.917 & \underline{33.21} \\
ControlNeXt \cite{peng2024ControlNeXt}  & 17.81 & 0.758 & 0.119 & 0.910 & 88.64 & 17.45 & 0.643 & 0.125 & 0.865 & 138.42 \\
\midrule
\rowcolor{rowcolor}
M2HVideo (\textbf{ours}) &\textbf{23.47}& \textbf{0.886} &\textbf{0.069} & \textbf{0.935} & \textbf{13.35} & \textbf{25.23} & \textbf{0.891} & \textbf{0.041} & \textbf{0.938} & \textbf{24.86} \\
\bottomrule
\end{tabular}
}
\end{table*}

\subsubsection{Evaluation Metrics}
As described in Section~\ref{method:sec3.2}, we evaluate our framework and the baselines across three dimensions: (i) clothing consistency, assessed using peak signal-to-noise ratio (PSNR), structural similarity (SSIM), and learned perceptual image patch similarity (LPIPS). These metrics are computed between the generated and ground-truth videos within the clothing regions, where higher PSNR and SSIM values indicate greater similarity, and lower LPIPS values correspond to improved perceptual fidelity; (ii) identity preservation, measured by cosine similarity (CSIM) in the facial embedding space, where higher values signify better preservation of the identity; and (iii) visual-temporal fidelity, evaluated using the Fr\'echet video distance (FVD), which quantifies the temporal coherence by comparing the distribution of generated videos with that of the ground-truth sequences. Lower FVD values imply better alignment with the temporal dynamics of the ground-truth data.

\subsubsection{Qualitative Comparison}
We performed extensive qualitative comparisons of the proposed M2HVideo framework with five state-of-the-art baseline methods using three datasets: the UBC fashion dataset, the ASOS dataset, and the MannequinVideos dataset. The visual results are presented sequentially in Figs. \ref{fig:UBC}, \ref{fig:exp_ASOS}, and \ref{fig:exp_mann}.
On the UBC fashion dataset, shown in Fig. \ref{fig:UBC}, existing pose transfer methods can be seen to suffer from several limitations. For instance, MagicPose produces clothing that appears darker than the input and is poorly aligned with the subject. 
These artifacts can be attributed to the appearance control module in MagicPose, which jointly encodes clothing and head features, resulting in entangled representations that compromise visual fidelity.
While AnimateAnyone preserves facial features relatively well, it produces desaturated colors of the clothing. MimicMotion has improved consistency but introduces an unnatural terracotta hue in the central region of the clothing, particularly in the front part of the video, as shown in the first and second columns. 
As a virtual try-on method, StableVITON achieves better clothing consistency but fails to maintain facial detail, rendering the facial features barely discernible. 
The relatively poorer performance of StableVITON may be ascribed to its architectural design, which incorporates facial features only at the UNet decoder stage. This design limits the preservation of low-level visual features and results in suboptimal reconstruction of facial details.
ControlNeXt further compromises the quality of the representation of the clothing by failing to retain the original texture and color and incorrectly altering the background color from white to gray. 
The limited performance of ControlNeXt can be attributed to its reliance on a single control signal. When multiple conditions are concatenated and processed by its lightweight encoder, the model struggles to disentangle them effectively. This leads to interference between the control pathways and results in an inaccurate facial synthesis.
Figure \ref{fig:exp_ASOS} shows the results on the ASOS dataset. In the video generated by MagicPose, both the color and the texture of the clothing fluctuate considerably from frame to frame.
MimicMotion has improved color fidelity relative to its performance on the UBC fashion dataset but introduces color shifts, such as turning a white bag yellow in its third and fourth columns of synthetic frames.
Both AnimateAnyone and ControlNeXt misidentify the front of a person as the back and alter the colors of the clothing——obvious errors. StableVITON continues to perform poorly at facial rendering. In contrast, our method consistently outperforms all the baselines by preserving fine-grained clothing attributes such as texture and color throughout the video frames, while maintaining high fidelity in the preservation of identity.
Finally, we evaluate our method on the MannequinVideos dataset, which includes professionally captured mannequin sequences. As illustrated in Fig. \ref{fig:exp_mann}, MimicMotion and ControlNeXt not only distort the colors but also fail to generate realistic feet for the mannequin. In comparison, our method effectively retains detailed clothing features, including accurate colors, fabric textures, and wrinkles.

\subsubsection{Quantitative Comparison}
For comparison, the results of the quantitative evaluation of all methods are presented in Table \ref{Quantitative_UBC_ASOS_fashion}, using the UBC fashion and ASOS datasets. Both datasets contain paired data suitable for an objective evaluation.
As outlined in Section \ref{method:sec3.2}, the evaluation covers three key aspects: clothing consistency, identity preservation, and visual-temporal fidelity.
In terms of clothing consistency, M2HVideo outperforms all the other methods in terms of all three metrics (PSNR, SSIM, and LPIPS) on both datasets.
On the UBC fashion dataset, it attains a PSNR of 23.47, slightly outperforming StableVITON. The SSIM score of 0.886 further establishes the structural accuracy of M2HVideo, exceeding the second-best value of 0.878 from MimicMotion. M2HVideo achieves the lowest LPIPS score of 0.069, which is slightly lower than that of MimicMotion. On the ASOS dataset, M2HVideo obtains a PSNR of 25.23 and an SSIM of 0.891, outperforming StableVITON. The LPIPS score drops to 0.041, clearly outperforming the second-best result, which is from MimicMotion. The preservation of facial identity by M2HVideo, in terms of CSIM, is consistently superior on both datasets. On the UBC fashion dataset, M2HVideo achieves a CSIM of 0.935, marginally surpassing AnimateAnyone, which is designed with identity-preserving modules. Other baselines have lower values of the CSIM, with MimicMotion and StableVITON scoring 0.928 and 0.912, respectively. On the ASOS dataset, M2HVideo again achieves the highest CSIM, 0.938, followed by StableVITON and MimicMotion, while ControlNeXt achieves the lowest.  M2HVideo demonstrates clear advantages in visual-temporal fidelity, as measured by FVD. It achieves the lowest FVD on both datasets, 13.35 on UBC and 24.86 on ASOS, indicating better visual-temporal fidelity than all the competing methods. The next best-performing method, StableVITON, achieves FVD values of 31.06 and 33.21, respectively, while the other baselines, such as AnimateAnyone and MagicPose, achieve substantially higher FVD. Across both datasets and all evaluation dimensions, M2HVideo consistently outperforms the state-of-the-art baselines.

\begin{table*}[ht]
\caption{User study results showing the proportion of votes preferring M2HVideo (ours) over each baseline.
\textcolor{gray}{Gray} indicates the standard deviation of preference rates}
\label{tab:user_study}
\centering
\begin{tabular}{llccccc}
\toprule
\multirow{2}{*}{Dataset}& \multirow{2}{*}{Metric} & \multicolumn{5}{c}{Preference for M2HVideo over baseline in the user study (\%)} \\
\cmidrule(lr){3-7}
 & & MagicPose & MimicMotion & AnimateAnyone & StableVITON & ControlNeXt \\
\midrule
\multirow{4}{*}{UBC fashion} 
& Clothing consistency      & $97.2 \textcolor{gray}{\pm 1.65}$  & $90.8 \textcolor{gray}{\pm 1.43}$ & $96.6 \textcolor{gray}{\pm 1.25}$ & $84.4 \textcolor{gray}{\pm 2.83}$ & $97.4 \textcolor{gray}{\pm 1.43}$\\
& Identity preservation  & $97.8 \textcolor{gray}{\pm 1.10}$ & $92.8 \textcolor{gray}{\pm 1.52}$ & $75.9 \textcolor{gray}{\pm 3.64}$ & $91.6 \textcolor{gray}{\pm 1.03}$ & $94.4 \textcolor{gray}{\pm 1.06}$ \\
& Video fluency          & $98.0 \textcolor{gray}{\pm 1.05}$ & $86.0 \textcolor{gray}{\pm 1.56}$ & $83.4 \textcolor{gray}{\pm 2.26}$ & $90.6 \textcolor{gray}{\pm 1.89}$ & $86.0 \textcolor{gray}{\pm 0.56}$ \\
& Overall visual quality & $97.8 \textcolor{gray}{\pm 1.10}$ & $85.0 \textcolor{gray}{\pm 1.78}$ & $96.0 \textcolor{gray}{\pm 1.33}$ & $93.6 \textcolor{gray}{\pm 2.57}$ & $96.4 \textcolor{gray}{\pm 1.62}$ \\
\midrule
\multirow{4}{*}{ASOS} 
& Clothing consistency      & $98.2 \textcolor{gray}{\pm 2.23}$ & $87.2 \textcolor{gray}{\pm 1.34}$ & $98.2 \textcolor{gray}{\pm 0.99}$ & $77.0 \textcolor{gray}{\pm 4.03}$ & $94.8 \textcolor{gray}{\pm 1.43}$\\
& Identity preservation  & $99.0 \textcolor{gray}{\pm 0.71}$ & $86.0 \textcolor{gray}{\pm 1.56}$ & $93.0 \textcolor{gray}{\pm 1.78}$ & $87.2 \textcolor{gray}{\pm 1.71}$ & $98.6 \textcolor{gray}{\pm 0.95}$ \\
& Video fluency          & $99.2 \textcolor{gray}{\pm 1.26}$ & $85.2 \textcolor{gray}{\pm 1.17}$ & $88.2 \textcolor{gray}{\pm 1.60}$ & $90.0 \textcolor{gray}{\pm 1.56}$ & $94.6 \textcolor{gray}{\pm 0.82}$ \\
& Overall visual quality & $99.6 \textcolor{gray}{\pm 0.63}$ & $90.4 \textcolor{gray}{\pm 1.23}$ & $97.2 \textcolor{gray}{\pm 0.84}$ & $92.6 \textcolor{gray}{\pm 0.67}$ & $99.0 \textcolor{gray}{\pm 0.71}$ \\
\midrule
\multirow{4}{*}{MannequinVideos} 
& Clothing consistency      & $97.5 \textcolor{gray}{\pm 1.07}$ & $84.6 \textcolor{gray}{\pm 0.82}$ & $92.1 \textcolor{gray}{\pm 0.88}$ & $67.0 \textcolor{gray}{\pm 0.74}$ & $98.3 \textcolor{gray}{\pm 0.70}$ \\
& Identity preservation  & $96.7 \textcolor{gray}{\pm 1.23}$ & $80.0 \textcolor{gray}{\pm 0.92}$ & $67.0 \textcolor{gray}{\pm 0.57}$ & $88.3 \textcolor{gray}{\pm 0.79}$ & $87.5 \textcolor{gray}{\pm 0.82}$ \\
& Video fluency          & $99.6 \textcolor{gray}{\pm 0.32}$ & $76.7 \textcolor{gray}{\pm 0.52}$ & $80.0 \textcolor{gray}{\pm 0.92}$ & $94.1 \textcolor{gray}{\pm 1.35}$ & $85.0 \textcolor{gray}{\pm 0.97}$ \\
& Overall visual quality & $98.3 \textcolor{gray}{\pm 0.70}$ & $98.2 \textcolor{gray}{\pm 21.4}$ & $92.9 \textcolor{gray}{\pm 0.82}$ & $92.5 \textcolor{gray}{\pm 0.92}$ & $95.4 \textcolor{gray}{\pm 0.32}$ \\
\bottomrule
\end{tabular}
\end{table*}

\subsubsection{User Study}
In addition to the qualitative and quantitative evaluations, we conducted a user study to further assess the preservation of the details of the clothing and perceptual realism in videos generated by our M2H framework. The study involved three datasets: UBC fashion, ASOS, and MannequinVideos. For UBC fashion and ASOS, we randomly selected 50 videos from each dataset, along with corresponding clothing and identity images as input. For MannequinVideos, we used 12 videos covering four poses, each with three types of outfit (short-sleeved T-shirt, long-sleeved shirt, and one-piece dress). Each mannequin video was paired with a randomly selected identity image from UBC fashion or ASOS for evaluation.
The user study was designed to evaluate our method in terms of clothing consistency, identity preservation, video fluency, and overall visual quality. A total of ten participants, all unaffiliated with the authors, took part in the study, casting a total of 24,800 votes. Each participant was shown three videos and one image: the input clothing video and face image, a real-person video generated by our method, and a real-person video generated by a baseline method. The results, summarized in Table \ref{tab:user_study}, report the percentages of participants who preferred the videos generated by M2HVideo over those produced by the baselines. In particular, our approach shows clear advantages in preserving clothing details, maintaining facial identity, ensuring temporal fluency, and enhancing the overall perceptual quality. Our method is strongly preferred to MagicPose and ControlNeXt, with preference rates exceeding 90\% in terms of all evaluation metrics and datasets. Notably, the identity preservation of AnimateAnyone on the MannequinVideos dataset was rated at only 67\%. Upon analyzing the corresponding results, we attribute this to AnimateAnyone’s strategy of encoding identity information by duplicating the entire UNet and injecting the resulting embedding into the backbone UNet. While this design effectively preserves the facial identity in the generated videos, it struggles to maintain accurate clothing colors. As a result, M2HVideo still achieved a high preference rate of 92.9\% for overall visual quality.

\begin{table}[t]
\centering
\caption{Ablation study on the UBC fashion dataset, demonstrating the effects of  $\mathcal{H}$, $\mathcal{A}$, and the mirror loss on identity preservation and video quality; the best and second-best scores are marked in \textbf{bold} and \underline{underline}, respectively}
\label{tab:ablation_module}
\setlength{\tabcolsep}{4pt}
\resizebox{0.46\textwidth}{!}{
\begin{tabular}{cccccccc}
\toprule
$\mathcal{H}$ & $\mathcal{A}$&  $\mathcal{L}_{\text{mir}}$ & PSNR $(\uparrow)$ & SSIM $(\uparrow)$ & LPIPS$(\downarrow)$  & CSIM $(\uparrow)$ & FVD $(\downarrow)$\\
\midrule
$\times$ & $\times$ & $\times$ & 20.98 & 0.843 & 0.082 & 0.918 & 19.34 \\
$\checkmark$ & $\times$ & $\times$ & 23.20 & 0.875 & 0.076 & 0.926 & 15.20 \\
$\times$ & $\checkmark$ & $\times$ & 21.97 & 0.866 & 0.074 & 0.920 &14.80 \\
$\times$ & $\times$ & $\checkmark$ & 22.41 & 0.871 & 0.078 & 0.926 &15.62 \\
$\checkmark$ & $\checkmark$ & $\times$ & 23.00 & 0.879 & 0.072 & 0.923 &\underline{13.41} \\
$\checkmark$ & $\times$ & $\checkmark$ & 22.15 & 0.874 & 0.071 & \underline{0.933} &13.62 \\
$\times$ & $\checkmark$ & $\checkmark$ & \underline{23.34} & \underline{0.882} & \textbf{0.068} & 0.928 &14.00 \\
$\checkmark$ & $\checkmark$ & $\checkmark$ & \textbf{23.47} & \textbf{0.886} & \underline{0.069} & \textbf{0.935} & \textbf{13.35} \\
\bottomrule
\end{tabular}
}
\end{table}

\begin{figure}[t]
\centering
\includegraphics[width=0.5\textwidth]{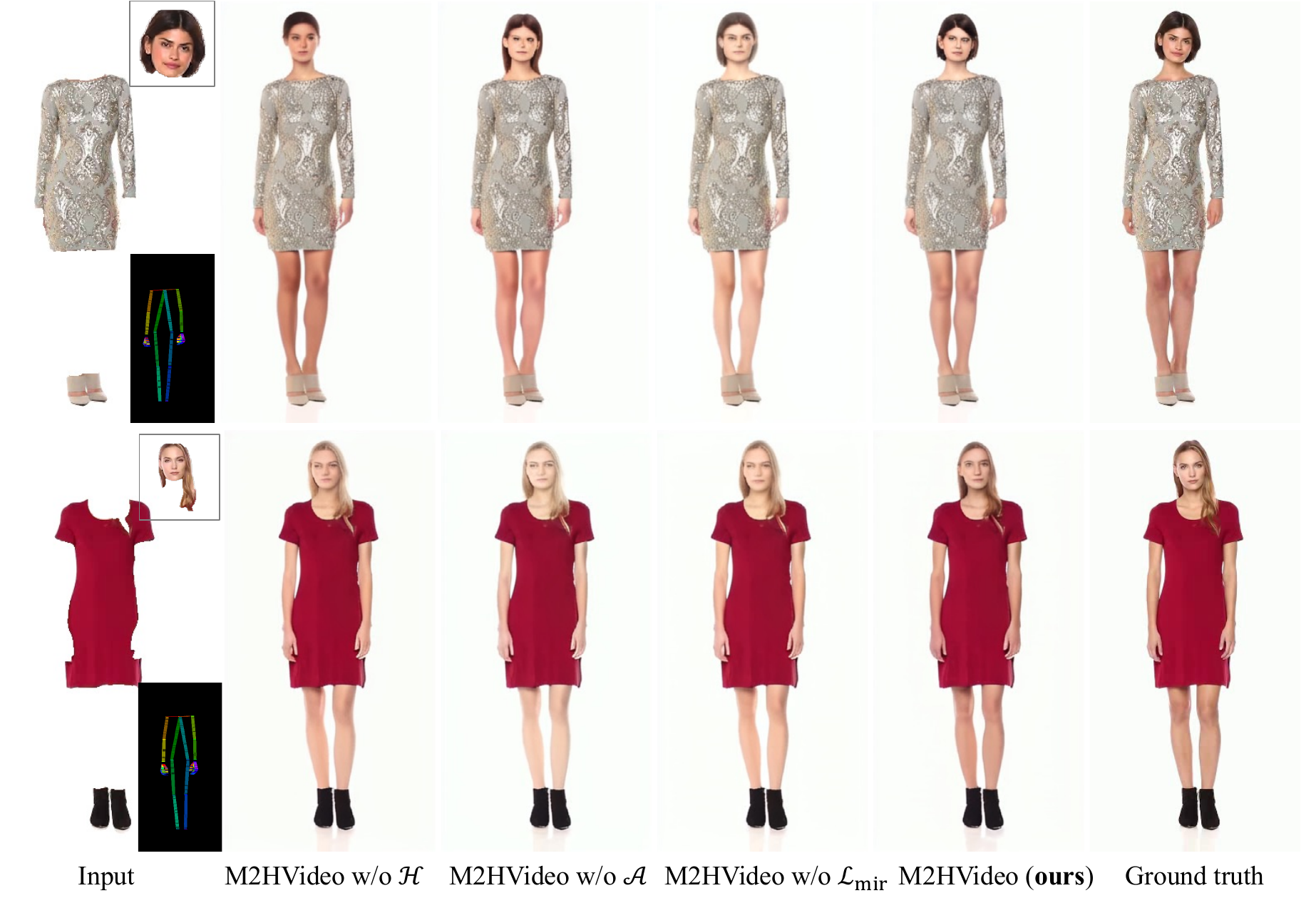}
\caption{Visual comparison of results from different variants of the M2HVideo, demonstrating the effects of including or excluding $\mathcal{H}$, $\mathcal{A}$, and $\mathcal{L}_{\text{mir}}$.}
\label{fig:exp_abl_module}
\end{figure}

\subsection{Ablation Study}
\label{exp:ablation}
This subsection presents the results of a series of ablation studies on the UBC fashion dataset to evaluate the individual contributions of three core components in the proposed M2HVideo framework: the dynamic pose-aware head encoder, the distribution-aware head-clothing adapter, and the customized training loss.

\subsubsection{Effectiveness of the Dynamic Pose-Aware Head Encoder}
To assess the effectiveness of the proposed dynamic pose-aware head encoder, we conducted an ablation study by removing its cross-attention layer in $\mathcal{H}$ and instead adding element-wise the head and body embeddings before inputting them into the UNet. The quantitative results are presented in Table \ref{tab:ablation_module}. As shown, excluding the dynamic pose-aware head encoder results in a noticeable degradation in the quality of the generated images, particularly in terms of CSIM. For instance, comparisons between the first and second rows, third and fifth rows, fourth and sixth rows, and seventh and eighth rows in the table demonstrate consistent decreases in CSIM when the encoder is removed (with all other modules remaining). Moreover, qualitative examples in Fig. \ref{fig:exp_abl_module} further support this observation: the first row, second column shows substantial differences in the head position and facial structure compared to the ground truth, while the second row, second column shows clear deviations in the shape of the face. These findings underscore the importance of an explicit pose-head alignment for preserving facial consistency and enhancing the overall fidelity of the video.

\subsubsection{Effectiveness of the Distribution-Aware Head-Clothing Adapter}
An ablation study was conducted to evaluate the effectiveness of the proposed distribution-aware head-clothing adapter $\mathcal{A}$.
As shown in Table \ref{tab:ablation_module}, comparisons between the first and third rows, the second and fifth rows, the fourth and seventh rows, and the sixth and eighth rows indicate that incorporating the adapter consistently improves the FVD score. This improvement mitigates the distributional shifts introduced by the temporal layers.
Visual comparisons in Fig. \ref{fig:exp_abl_module} further illustrate the benefit. Without the adapter, notable distortions occur in the distribution of the facial features and at the junction between the head and clothing.
For instance, in the first row, the region where the clothing contacts the head is erroneously transformed into long hair. These results suggest that the adapter $\mathcal{A}$ enhances both the visual accuracy and the spatial coherence in the generated videos.

\begin{table}[t]
\centering
\caption{Parameter sensitivity study results on the UBC Fashion Dataset, illustrating the effect of varying $\alpha$ and $\beta$ in the mirror loss on identity preservation and video quality; the best and second-best scores are marked in \textbf{bold} and \underline{underline}, respectively}
\label{tab:ablation_results_mirror}
\setlength{\tabcolsep}{4pt}
\resizebox{0.46\textwidth}{!}{
\begin{tabular}{ccccccc}
\toprule
$\alpha$ & $\beta$ & PSNR $(\uparrow)$ & SSIM $(\uparrow)$ &LPIPS $(\downarrow)$& CSIM $(\uparrow)$ & FVD $(\downarrow)$\\
\midrule
0 & 0.005 & \underline{23.35} & 0.865 & 0.073 & 0.928 &14.58 \\
0.005 & 0.005 & 23.31 & 0.878 & \textbf{0.068} & 0.930 &14.72 \\
0.05 & 0.005 & \textbf{23.47} & \textbf{0.886} & \underline{0.069} & \textbf{0.935} & \textbf{13.35} \\
0.5 & 0.005 & 22.56 & 0.874 & 0.075 & \underline{0.933} & \underline{13.66} \\
0.05 & 0 & 20.41 & 0.842 & 0.101 &0.907 &30.37 \\
0.05 & 0.0005 & 22.17 & \underline{0.879} & 0.071 & 0.929 & 13.78 \\
0.05 & 0.05 & 20.6 & 0.843 & 0.102 &0.921 &23.31 \\
0.05 & 0.5 & 18.04 & 0.801 & 0.149 &0.902 &27.36 \\
\bottomrule
\end{tabular}
}
\end{table}

\subsubsection{Effectiveness of the Mirror Loss}
We conducted ablation studies to evaluate the effectiveness of the proposed mirror loss. Table \ref{tab:ablation_module}, particularly the fifth and eighth rows, shows that incorporating the mirror loss consistently improves both the overall quality of the generated video and the fidelity of the facial details. Similarly, the fourth column of Fig. \ref{fig:exp_abl_module} demonstrates that the addition of the mirror loss yields clearer hair textures and more distinct facial features compared to the setting without it.
To examine the influence of the terms in the mirror loss, we varied $\alpha$ and $\beta$. The results are presented in Table \ref{tab:ablation_results_mirror}.
The best performance is achieved when $\alpha$ is set to 0.05 and $\beta$ to 0.005. Setting either term to zero causes substantial degradation; for instance, $\beta$ set to 0 yields an FVD of 30.37 and an LPIPS of 0.101, while setting $\alpha$ to 0 reduces the CSIM to 0.928.
Excessively large weights, such as an $\alpha$ of 0.5 or a $\beta$ of 0.5, also harm the performance due to over-regularization, as indicated by the increased values of LPIPS and FVD.
These results confirm that balanced mirror loss terms are essential for preserving visual fidelity, identity, and temporal consistency.

\section{Conclusion}
\label{conclusion}
This paper presented M2HVideo, a framework for converting mannequin-based clothing displays into photorealistic human videos while preserving clothing fidelity and enabling identity control of avatars. The method addresses key limitations of conventional presentation formats in online fashion media. To ensure the consistency of the clothing and the preservation of identity, we introduced a dynamic pose-aware head encoder and a distribution-aware adapter. These modules enable precise estimation of the facial pose and reduce the shifts in the distributions during generation. In addition, we proposed a mirror loss in pixel-space reconstruction to recover high-frequency facial details lost during latent compression. Extensive experiments on three datasets, UBC fashion, ASOS, and MannequinVideos, show that M2HVideo outperforms existing state-of-the-art methods in the consistency of the clothing, the preservation of the identity, and the fidelity of the video.

\bibliographystyle{IEEEtran}
\bibliography{references}

\vfill
\end{document}